%% file: main.tex
\documentclass[11pt]{article}

\usepackage[preprint]{acl}

\usepackage{common}
\usepackage{siunitx}
\usepackage{microtype}

\usepackage{fontspec}
\usepackage{xeCJK}
\setCJKmonofont{FandolSong-Regular.otf}

\tcbset{colback=gray!5!white, colframe=gray!75!black, fonttitle=\bfseries, width=\linewidth}

\newcommand{\contentfuzz}{\textsc{ContentFuzz}\xspace}
\newcommand{\favor}{\texttt{Favor}\xspace}
\newcommand{\against}{\texttt{Against}\xspace}
\newcommand{\neutral}{\texttt{Neutral}\xspace}
\newcommand{\logprobs}{\texttt{logprobs}\xspace}

\title{Content Fuzzing for Escaping Information Cocoons on Social Media}

\author{
    Yifeng He\textsuperscript{\rm 1} \and
    Ziye Tang\textsuperscript{\rm 2} \and
    Hao Chen\textsuperscript{\rm 3} \\
    \textsuperscript{\rm 1}Department of Computer Science, 
    \textsuperscript{\rm 2}Department of Communication \\
    University of California, Davis \\
    \textsuperscript{\rm 3}University of Hong Kong \\
    {\{yfhe,\, szytang\}@ucdavis.edu,\, chenho@hku.hk}
}

\begin{document}

\maketitle

\input{src/abstract.tex}
\input{src/introduction.tex}
\input{src/background.tex}

\input{src/design.tex}

\input{src/experimental_setup.tex}
\input{src/results.tex}
\input{src/conclusion.tex}

\input{src/limitations.tex}
\input{src/ethics.tex}

\section*{Acknowledgments}
This material is based upon work supported by UC Noyce Initiative.

\bibliography{main}

\input{src/appendix.tex}

\end{document}

%% file: src/abstract.tex
\begin{abstract}
	Information cocoons on social media limit users' exposure to posts with diverse viewpoints.
	Modern platforms use stance detection as an important signal in recommendation and ranking pipelines,
	which can route posts primarily to like-minded audiences and reduce cross-cutting exposure.
	This restricts the reach of dissenting opinions and hinders constructive discourse.
	We take the creator's perspective and investigate how content can be revised to reach beyond existing affinity clusters.
	We present \contentfuzz,
	a confidence-guided fuzzing framework that rewrites posts while preserving their human-interpreted intent and induces different machine-inferred stance labels.
	\contentfuzz aims to route posts beyond their original cocoons.
	Our method guides a large language model (LLM) to generate meaning-preserving rewrites using
	confidence feedback from stance detection models.
	Evaluated on four representative stance detection models across three datasets in two languages,
	\contentfuzz effectively changes machine-classified stance labels,
	while maintaining semantic integrity with respect to the original content.
\end{abstract}

%% file: src/introduction.tex
\section{Introduction}

Social media platforms increasingly mediate how people access information.
However, selective exposure often confines users to highly homogeneous content environments,
known as information cocoons~\cite{he2023informationcocoons,zhou2025impact}.
Such confinement narrows individuals' perspectives and limits informational diversity.
Moreover, information cocoons can hinder intellectual growth, reinforce social segmentation,
and contribute to emotional or psychological harm~\citep{Simpson2017skinny,NapoliDwyer2018usmediapolicy}.
These cocoons often arise because social media platforms restrict the range of viewpoints that users encounter.
Users naturally gravitate toward content sharing a similar stance on various topics,
especially early in their interactions with a platform.
While this helps recommender systems learn user preferences and deliver personalized content to increase engagement~\cite{he2022bigdata},
it also creates a feedback loop in which users are repeatedly exposed to content aligned with their existing beliefs, deepening the cocoon over time.

Information cocoons affect both users and content creators.
For users, confinement to homogeneous information environments limits exposure to diverse viewpoints,
reinforcing existing biases and amplifying ideological polarization~\citep{garimella2018political}.
Cross-cutting exposure supports healthier deliberation and reduces affective polarization,
while homogeneous and repetitive content can intensify emotional stress and reduce well-being~\citep{Alzeer2017cocoons}.
For content creators and publishers, these dynamics impose practical constraints:
posts circulate primarily within affinity clusters,
making it difficult for high-quality content to reach broader or cross-cutting audiences.
Consequently, enabling content to escape information cocoons is important for both improving users' informational diversity and enhancing the visibility of creators' messages.

Despite this importance, escaping information cocoons remains technically challenging.
Recommendation pipelines and ranking mechanisms operate as black boxes,
making it difficult to determine how subtle, semantics-preserving edits influence a post's exposure.
Even minor phrasing changes can shift downstream model behavior~\cite{zhang2025steerdiff,he2025tfbench},
yet existing research largely attempts to mitigate cocoon effects through platform-side algorithmic interventions~\cite{krause2024mitigatingexposurebias,li2025contentagnostic}.
\citet{ma2025breakinginformationcocoons}, for example,
analyze how diversity-oriented ranking and re-ranking algorithms affect homogenization dynamics and propose algorithmic adjustments to mitigate these effects.

While such algorithmic interventions provide valuable insights into cocoon formation,
they remain fundamentally platform-controlled.
Individual users and content creators cannot modify recommender algorithms,
nor do they have visibility into how their posts are filtered, ranked, or delivered.
As a result, users and creators have limited agency to expand content reach beyond existing affinity clusters.
This gap motivates content-wise approaches that operate independently of platform algorithms.
From the creator's perspective, \emph{escaping information cocoons},
achieved by identifying semantic-preserving rewrites that keep a post's human-interpreted stance but change a stance analyzer's predicted label,
offers a practical mechanism for broadening cross-group exposure without relying on opaque platform-side changes.

\begin{figure*}[t]
	\includegraphics[width=1\linewidth]{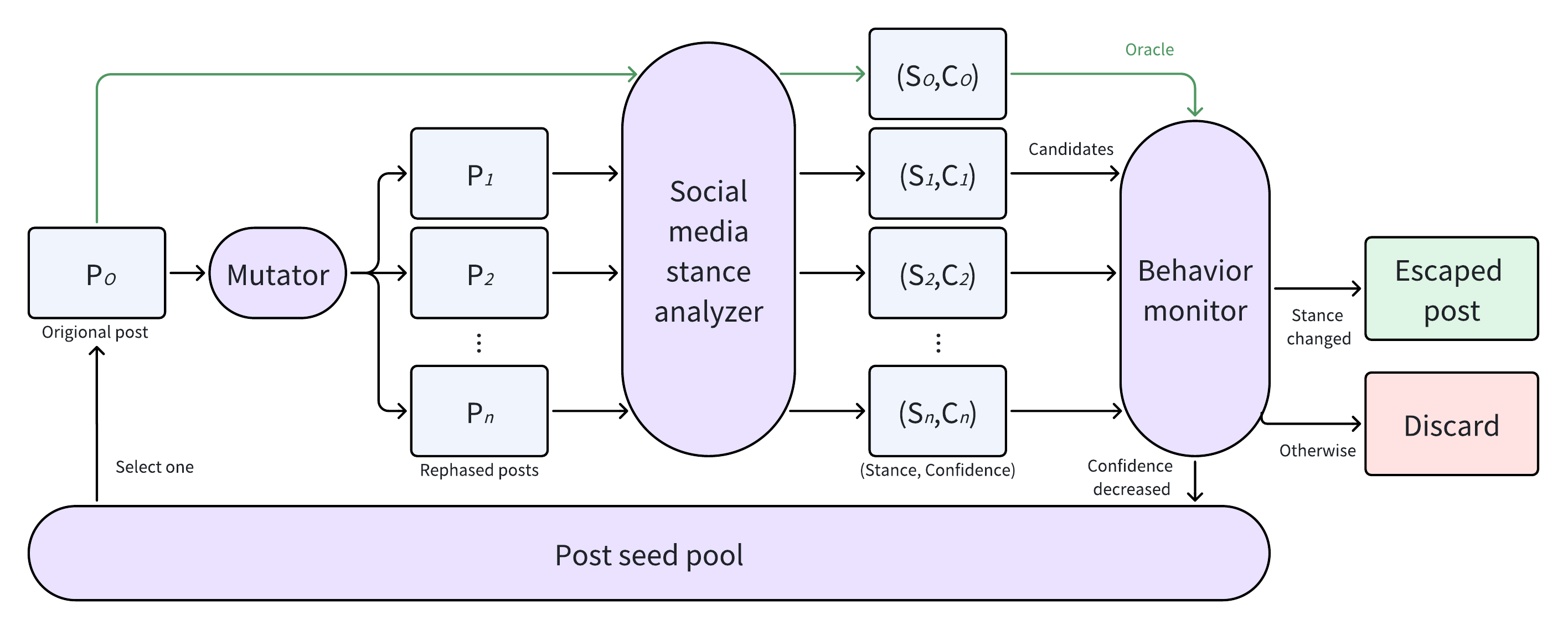}
	\caption{Post content generation in \contentfuzz. \emph{Seed} denotes candidate posts stored for mutation.}
	\label{fig:workflow}
\end{figure*}

We introduce \contentfuzz,
a novel automated content-wise framework to mitigate information cocoons.
Our approach targets stance detection models,
which constitute a core signal in social media recommendation pipelines for assessing ideological orientation and structuring public-opinion discourse on contentious topics~\cite{Atske_2019,zhang2024surveystance,Muthusami2025InterpretableStance}.
We adapt fuzzing, a methodology from software testing, to iteratively discover such rewrites.
Inspired by recent advances in LLM jailbreak fuzzing~\cite{yu2024llmfuzzer,liu2024autodan},
\contentfuzz leverages confidence of the stance analysis as feedback to guide a generative LLM in producing semantic-preserving rewrites.
Through this feedback-guided process,
\contentfuzz reliably alters machine-classified stance labels while preserving the post's human-interpreted stance,
thereby enabling content to reach audiences outside its existing cocoon.
\contentfuzz is model-agnostic, cross-lingual, cross-topic, and readily adaptable to a wide range of social media scenarios.
In our experiments across three real-world datasets in two major languages and four stance detection models,
\contentfuzz consistently enables posts to escape information cocoons
with robust semantic integrity and fluency in the generated rewrites.
To the best of our knowledge,
\contentfuzz is the first content-side computational approach toward mitigating information cocoon effects.

%% file: src/background.tex
\section{Background and related work}

\subsection{Information cocoons}

Information cocoons arise when algorithmic curation and selective exposure confine users
to homogeneous content environments,
limiting informational diversity and reinforcing existing beliefs~\cite{he2023informationcocoons,zhou2025impact}.
Cocoon effects have been documented in news sharing~\cite{du2024newscocoons},
video platforms~\cite{yi2023cocoons},
and social media~\cite{chen2025cocoonparadox,wang2025informationcocoons},
with \citet{piao2023humanai} attributing their emergence to human--AI adaptive dynamics.
Current mitigation strategies are platform-controlled,
such as diversity-oriented re-ranking~\cite{ma2025breakinginformationcocoons},
and are therefore unavailable to content creators.
\contentfuzz explores a complementary, creator-side direction:
searching for semantics-preserving rewrites that shift a post's machine-classified stance label.
Such shifts probe whether content can cross stance-conditioned filtering boundaries
without altering its human-interpreted meaning.
We present this search as an iterative, feedback-guided process
built on techniques from software fuzzing.

\subsection{Fuzzing}

Fuzzing is the process of dynamically testing software by iteratively generating random inputs~\cite{miller1990empirical}.
Modern software testing widely adopts gray-box coverage-guided fuzzing,
which leverages code coverage as feedback to direct the input generation process~\cite{bohme2016coverage,bohme2017directed}.
Some recent work also explores applying fuzzing to augment language models~\cite{zhao2023understanding,huang2024code,he2025fuzzaug}.
At a high level, fuzzing consists of three core components: %
iterative input generation, feedback-based selection, and seed scheduling~\cite{Zeller2019fuzzingbook}.

\paragraph{Input generation}
Fuzzing begins from one or more seed inputs and iteratively produces variants through mutation.
Rather than exhaustively enumerating all possible inputs,
fuzzing aims to efficiently discover transformations that induce new or interesting behaviors~\cite{chen2018angora,chen2019matryoshka}.
Fuzzers can also generate structured inputs,
extending to various software domains~\cite{zhang2024llamafuzz,rong2025irfuzzer,tu2026cottontail}.

\paragraph{Feedback-guided selection}
After generating a new input,
the fuzzer executes the target system and observes its behavior.
In gray-box settings,
this signal only needs to correlate with progress toward a desired outcome~\cite{bohme2016coverage,rong2022valkyrie}.
Inputs triggering interesting new behaviors are retained as seeds for future iterations.

\paragraph{Seed scheduling}
Given a pool of candidate seeds,
fuzzers prioritize which inputs to mutate next based on their historical performance.
Seed scheduling estimates a seed's potential to yield bugs,
enabling the fuzzer to focus on promising regions of the input space~\cite{woo2013scheduling,xu2024graphuzz}.
This prioritization is critical for efficiency when compute hours are limited.

\paragraph{Fuzzing for language models}
Fuzzing has also been applied to neural networks, including large language models (LLMs).
Bugs in LLMs include jailbreaks,
which make the models generate harmful, biased, or toxic content~\cite{perez2022red,chao2025jailbreaking}.
Prior work applies fuzzing-style search to mutate jailbreak prompt templates,
using learned classifiers or likelihood-based fitness functions as feedback signals to guide mutation~\cite{yu2024llmfuzzer,liu2024autodan}.
Fuzzing LLMs requires novel designs for input generation, behavior monitoring, and seed scheduling,
since LLMs differ significantly from traditional software.

\subsection{Stance detection}

Stance detection, also referred to as stance analysis, is a natural language classification task
that aims to identify the stance or attitude of the author expressed in a piece of text towards a specific target or topic~\cite{mohammad2016sem16,zhang2024surveystance}.
Stance detection is often used in social media analysis and recommendation,
where the platforms expose users to content strongly aligned with their own side with stance-conditioned feed ranking~\cite{garimella2018political,Aldayel2019Your,li2025contentagnostic}.
Modern approaches to stance detection use
fine-tuned embedding models~\cite{liu2021enhancing,conforti2020wtwt,allaway2020vast,liang2022zeroshot,ding2025zeroshot}
and generative models~\cite{li2023stance,taranukhin2024stance-reasoner,gatto2023chain,zhao2024zerostance,lan2024cola}.
In this work, we focus on post-level stance detection.

\subsection{LLM-based text rewriting}

LLM-based text rewriting has recently been applied to social media content
with objectives ranging from content moderation to engagement optimization.
\citet{ziegenbein2024rewriting} use reinforcement learning from machine feedback
to rewrite inappropriate argumentation while preserving core claims.
\citet{pillai2025engagement} rewrite news tweets to control engagement properties,
while \citet{santos2025polarization} and \citet{wang2025toxicmitigation}
target polarization reduction and toxic language mitigation, respectively.
These methods optimize a social property of the text,
whether tone, engagement, or toxicity.
\contentfuzz pursues a different objective:
rewrites that flip a stance analyzer's predicted label
while preserving the post's semantic content.
The optimization target is the classifier's decision boundary rather than a social property of the text,
which aligns \contentfuzz with adversarial robustness testing for stance models.

%% file: src/design.tex
\section{Design}

In this section, we describe the design of \contentfuzz.
The workflow is depicted in \autoref{fig:workflow} and detailed in \autoref{alg:contentfuzz}:
starting from a single post as the seed,
\contentfuzz mutates it into candidates,
runs the stance analyzer to obtain a confidence score for each candidate,
keeps confidence-lowering candidates for future mutations,
and stops when a candidate changes the predicted stance or when iterations are exhausted.
We then detail the three key components:
feedback guidance (\autoref{sec:feedback}),
seed scheduling (\autoref{sec:seed-scheduling}),
and mutation of the selected seeds (\autoref{sec:mutation}).

\begin{algorithm}[t]
	\caption{Confidence-guided content fuzzing}
	\label{alg:contentfuzz}
	\small
	\input{fig/contentfuzz_algo.tex}
\end{algorithm}

\subsection{Feedback guidance} \label{sec:feedback}

\paragraph{Challenge}
Stance analyzers based on large language models (LLMs) operate as black boxes and are difficult to interpret~\cite{odena19tensorfuzz}.
Black-box testing such systems without feedback is inefficient~\cite{bohme2016coverage},
so defining an effective feedback mechanism to guide gray-box fuzzing for these systems is challenging.
Previous work~\cite{xie2019deephunter,odena19tensorfuzz} tracks internal neuron activations in deep neural networks as coverage metrics.
\citet{park2023gradfuzz} proposed gradient vector coverage, which leverages gradients obtained by partially differentiating the cross-entropy loss function as feedback.
These techniques are impractical in our setting because they require access to internal structures of the target model and do not scale to transformer-based LLMs with hundreds of millions or billions of parameters.

\subsubsection{Analysis confidence as feedback}
Let us reconsider our fuzzing objective.
Unlike previous work that seeks to find \emph{all} inputs that trigger unexpected behaviors,
we focus on identifying \emph{one} single variant of a given post that changes the target model's predicted stance.
Therefore, we do not aim for completeness in our guidance metric.
Instead, we require a metric that reliably indicates whether a mutated candidate is closer to escaping the original stance.
To this end, we use the \emph{analysis confidence score} returned by the target stance analyzer as our feedback metric.
We describe methodologies to obtain confidence scores from two types of stance analyzers:
fine-tuned encoder-based classifiers (\eg, BERT~\cite{devlin2019bert}) and generative LLMs (\eg, Gemini-2.5~\cite{comanici2025gemini25}).
In the following text,
we use $x = \{x_1, \ldots, x_n \}$ to denote the tokens of the input prompt,
which contains the post content, the target topic, and the instruction to generate a stance response if any.
We use $\theta$ to denote the parameters of the target stance analyzer.
Using confidence feedback does not require any instrumentation of the target model.

\paragraph{Classifier stance analyzers}
Fine-tuning stance analyzers typically adds a softmax-based classification head on top of a pre-trained masked language model~\cite{sun2019finetune}.
The encoder maps the input post to a vector representation,
which is fed into a linear classification layer to produce a logit $z_i$ for each label $k \in$ \{\favor, \against, \neutral\}.
The softmax layer then maps these logits into a probability distribution~\cite{bridle1989training,hinton2015distilling},
and $\hat{k}$ is the label with the highest probability:
\[
	P_{\theta}(k | x) = \frac{\exp(z_k)}{\sum_{j} \exp(z_j)},
	\hat{k} = \arg\max_{k} P_{\theta}(k | x).
\]
We use the probability of the predicted stance $\hat{k}$ as the analysis confidence score to guide fuzzing:
\begin{equation*}
	\confmlm(x, \hat{k}) = P_{\theta}(\hat{k} | x).
\end{equation*}

\paragraph{Generative stance analyzers}
Recent work has investigated using generative LLMs for stance analysis~\cite{lan2024cola}.
These approaches prompt the LLM with the post content and the target topic,
asking it to generate a response that indicates the stance.
Generative causal language models employ autoregressive decoding,
predicting one token at a time based on previously generated tokens and the input prompt~\cite{radford2019gpt2,brown2020gpt3}.
Each decoding step is a classification task over the vocabulary.
Consequently, each predicted token has an associated probability distribution,
often exposed as \logprobs by the LLM serving API.
\logprobs are the natural logarithms of the model-assigned probabilities for each token in the vocabulary and are often used as a measure of generation confidence to mitigate hallucinations~\cite{xu2025logprobs,zhang2025actlcd}.
Let $y = \{y_1, \ldots, y_m\}$ denote the tokens in the generated stance response.
Then the \logprobs for each generated token $y_i$ are
\[
	l_i = \log p_{\theta}(y_i | x, y_{<i}).
\]
Note that the joint probability of generating a sequence of tokens is the product of the conditional probabilities of generating each token~\cite{bengio2000neural,radford2019gpt2}:
\[
	p_{\theta}(y | x) = \prod_{i=1}^{m} p_{\theta}(y_i | x, y_{<i}).
\]
With the basic rules of logarithms, we have
\[
	L(x, y) =  \sum_{i=1}^{m} \log p_{\theta}(y_i | x, y_{<i}) = \sum_{i=1}^{m} l_i.
\]
Our feedback for fuzzing generative stance analyzers is the exponential of the joint \logprobs:
\begin{equation*}
	\confclm(x, y) = \exp(L(x, y)).
\end{equation*}

\subsection{Seed scheduling} \label{sec:seed-scheduling}
After adding the mutated candidates of interest to the seed pool,
\contentfuzz selects the next seed post to fuzz from the pool.
Our goal is to identify a mutated candidate that escapes the original stance in as few iterations as possible,
so we prioritize seeds that are more likely to lead to successful escapes.
These seeds have lower analysis-confidence scores,
indicating that they lie closer to the decision boundary of the target stance analyzer.
Motivated by this observation, we design our seed-scheduling strategy using a min-heap,
where \contentfuzz always selects the seed with the lowest confidence score in the entire pool for mutation.
We also present and discuss other seed-scheduling design choices in \autoref{sec:ablation:seed-scheduling},
where we compare the effectiveness and efficiency of alternative strategies.

\subsection{Mutation} \label{sec:mutation}

\subsubsection{LLM-based rewriting}

After selecting a seed post,
\contentfuzz mutates its content to generate new candidate posts.
To achieve this, we design an LLM-based mutator with a strict prompt dedicated solely to rewriting.
Unlike software fuzzing with multiple mutators~\cite{fioraldi2022libafl},
we enforce a single rewrite mutation in \contentfuzz to ensure the semantic integrity of the posts.
After selecting the seed, we wrap its content in templates (\appautoref{fig:rewrite_prompts_en})
and send it to an instruction-tuned LLM to produce mutated candidates.

To accelerate exploration and avoid frequent mutation failures,
which can terminate fuzzing at an early stage if the pool becomes depleted,
we allow the mutator to generate multiple candidates in a single mutation step.
In \contentfuzz, we let the mutator generate 5 candidates,
and evaluate each against the target stance analyzer individually.

In \contentfuzz, we use Gemini-2.5-Flash-Lite~\cite{comanici2025gemini25}.
The mutator performs constrained paraphrasing under a strict prompt template (\appautoref{fig:rewrite_prompts_en}),
a narrow task that does not require external knowledge or multi-step reasoning.
Recent work validates LLM-based rewriting in closely related settings~\cite{ziegenbein2024rewriting,pillai2025engagement}.
To maximize fuzzing throughput and minimize cost,
we choose a smaller and faster model to reduce the overhead of using an LLM in the fuzzing process while maintaining competitive performance.
To further increase fuzzing throughput,
we disable the model's chain-of-thoughts reasoning capability by setting the thinking-token budget to 0.

\subsubsection{Temperature scheduling}
Temperature in generative LLMs is commonly used to control the level of creativity
~\cite{xu2022codellm,peeperkorn2024temperature,renze2024effect,zhu2024adaptive}.
However, deciding on a fixed temperature for \contentfuzz is challenging.
Different social-media platforms and different topics may require different levels of creativity in rewriting.
Moreover, \contentfuzz implements only a single, strict mutation operator, for which a fixed temperature may lead to suboptimal exploration-exploitation trade-offs~\cite{bohme2017directed,rong2022valkyrie,luo2023selectfuzz}.
To address these challenges, we propose \emph{temperature scheduling}, which dynamically adjusts the temperature during fuzzing.

We discretize the range of temperatures~\cite{gemini-doc}
into a finite set $\mathcal{T} = \{0.0, 0.1, \dots, 2.0\}$.
For each temperature $t \in \mathcal{T}$,
we assign the initial energy value $E_t = 1.0$ for a uniform prior sampling probability.
At each fuzzing iteration,
we randomly select a temperature $t$ from $\mathcal{T}$ with probability
\[
	P(t) = \frac{E_t}{\sum_{t' \in \mathcal{T}} E_{t'}}.
\]
Suppose the mutator generates $N$ candidates using temperature $t$,
and $s$ of them successfully reduce the analysis confidence compared with their parent seed.
We update the energy of $t$ by the mutation success rate of the current iteration
\[
	E_t \gets E_t + \frac{s}{N}.
\]
This adaptive scheduling allows \contentfuzz to dynamically select temperatures that have historically produced higher-quality variations.
With temperature scheduling,
\contentfuzz seamlessly generalizes across social-media platforms, topics, and target stance analyzers without manual tuning.

%% file: fig/contentfuzz_algo.tex
\begin{algorithmic}[1]
	\Function{\contentfuzz}{$post$, $N$}
	\\\Comment{$post =$ original post to apply \contentfuzz}
	\\\Comment{$N =$ number of allowed iterations for fuzzing}
    \State \textproc{Scheduler.Add}($post, 1.0$) 
	\For{$i = 1$ to $N$}
	    \State $seed \gets $ \textproc{Scheduler.Select}()
	    \State $mutants \gets$
        \Statex \hskip\algorithmicindent \textproc{Mutator.Rewrite}($seed.content$)
        \State $n\_succ \gets 0$
	    \ForAll{$m \in mutants$}
	        \State $stance, conf \gets$ \textproc{Analyze}($m$)
	        \If{$stance \neq seed.stance$}
	            \State \Return $m$ \Comment{Return successful escape}
	        \EndIf
            \If{$conf < seed.conf$}
                \State \textproc{Scheduler.Add}($m, conf$)
                \State $n\_succ \gets n\_succ + 1$
            \EndIf
	    \EndFor
        \State \textproc{Mutator.UpdateEnergy}($n\_succ$, $|mutants|$)
	\EndFor
	\State \Return Nothing \Comment{No escape found}
	\EndFunction
\end{algorithmic}

%% file: src/experimental_setup.tex
\section{Experimental setup} \label{sec:experimental-setup}

\subsection{Datasets}
We conduct experiments on three stance detection datasets spanning multiple social-media platforms and two languages:
SemEval2016-Task6 (Sem16), VAST, and C-STANCE.
Sem16~\cite{mohammad2016sem16} contains English tweets on six targets.
VAST~\cite{allaway2020vast} is collected from the Room for Debate section of the \textit{New York Times}
and contains English articles on \num{304} unique targets.
C-STANCE~\cite{zhao2023c-stance} is a Chinese dataset collected from Weibo with \num{48126} targets.
C-STANCE includes two subtasks: C-STANCE-A for target-based stance detection and C-STANCE-B for domain-based stance detection;
we use C-STANCE-A in our experiments.
All datasets are expert-annotated, 
each with a three-class stance scheme that we normalize to a unified set of labels:
\favor, \neutral, and \against.
Sem16 uses FAVOR/AGAINST/NONE;
VAST uses integer labels 0/1/2 (con/pro/neutral);
C-STANCE-A uses Chinese labels (支持/反对/中立).
All mappings are one-to-one with no granularity reduction.

\subsection{Targeted stance analyzers} \label{sec:stance-analyzers}

We evaluate \contentfuzz on three styles of stance analyzers: encoder-based models, zero-shot models,
and prompt-engineering models.
We select representative models for each style and describe their details in this section.
We provide the initial performance of these models in \appautoref{tab:performance}.

\paragraph{Encoder}
We use BERT~\cite{devlin2019bert} and RoBERTa~\cite{liu2019roberta} as our target encoder-based stance analyzers.
For English datasets Sem16 and VAST,
we use the released \texttt{bert-base-uncased} and \texttt{roberta-base} checkpoints.
For C-STANCE-A, we use the \texttt{chinese-bert-wwn} and \texttt{chinese-roberta-wwm-ext} checkpoints
pre-trained on Chinese corpora by \citet{cui2020revisiting,cui2021chinesebert}.
Details of these models are provided in \appautoref{appendix:fine-tuning}.

\paragraph{Zero-shot}
We use the term \emph{zero-shot} to refer to generative LLMs used without any parameter or architecture modification~\cite{radford2019gpt2,zhao2024zerostance},
\ie, without fine-tuning, prompt engineering, or in-context learning.
We directly prompt an LLM with the instructions in \appautoref{fig:cls_prompt} to perform stance detection.
We use Google Gemini-2.5-Flash-Lite~\cite{comanici2025gemini25} as our zero-shot model.
We set the temperature to 0 for more deterministic outputs and better reproducibility.
We apply guided decoding~\cite{scholak2021picard} to restrict the output space to valid stance only.

\paragraph{Prompt engineering}
Recent work also explores stance detection using tailored prompts~\cite{li2023stance,lan2024cola,taranukhin2024stance-reasoner}.
We evaluate our approach on COLA~\cite{lan2024cola},
the current state-of-the-art prompt-engineering method.
In COLA, LLMs are assigned distinct roles that form a collaborative system for analyzing stances in a given post.
We directly adapt the released open-source implementation of COLA.
We use Gemini-2.5-Flash-Lite for COLA.

\subsection{Research questions}

To validate the effectiveness of \contentfuzz,
we design experiments to answer the following research questions:
\begin{enumerate*}[label=\textbf{RQ\arabic*}]
	\item How effective is \contentfuzz across different stance analyzers and datasets?
	\item Do rewrites generated for one analyzer transfer to other unseen ones?
	\item How does temperature scheduling impact the effectiveness of \contentfuzz?
	\item How does seed scheduling influence the performance of \contentfuzz?
\end{enumerate*}

%% file: src/results.tex
\section{Evaluation results}

\subsection{Performance evaluation} \label{sec:performance-evaluation}

In this section, we evaluate the performance of \contentfuzz across three aspects:
success rate, semantic integrity, and fluency.
Success rate measures the effectiveness of \contentfuzz in rewriting posts to escape information cocoons,
while semantic integrity and fluency assess the quality of the generated rewrites by \contentfuzz.

\paragraph{Escape success rate}
We measure the escape success rate (ESR) as the percentage of posts
that are classified correctly by the targeted stance analyzer before fuzzing,
but are misclassified after being rewritten by \contentfuzz.
Let $D_{\text{corr}}$ denote the set correctly classified posts,
and let $CF$ denote the \contentfuzz function.
Then, for all $p \in D_{\text{corr}}$,
\[
	ESR = \frac{|\{p | p.stance \neq CF(p).stance\}|}{|D_{\text{corr}}|}.
\]

\paragraph{BERTScore}
BERTScore~\cite{Zhang2020BERTScore}
is a widely used metric to evaluate the semantic similarity between texts.
BERTScore uses an encoder model to compute contextual sentence embeddings for both the original post and the rewritten post.
We report the mean F1 score over successfully rewritten posts as the semantic integrity score.

\paragraph{Perplexity}
Perplexity (PPL)~\cite{jelinek1977perplexity} measures the model's uncertainty in predicting the next token in a sequence,
which provides a sense of fluency for generated text.
We follow AutoDAN~\cite{liu2024autodan} to report the perplexity of the rewritten posts.
Furthermore, we develop \emph{perplexity ratio} (PPLr) to measure the fluency of the generated rewrites
relative to their original posts.
Since absolute perplexity is sensitive to topic, style, and language-specific token distributions, directly comparing PPL values across different posts or languages can be misleading.
PPLr isolates the fluency change introduced by the rewriting process itself.
For each post $p$ that is successfully rewritten,
\[
	PPLr = \frac{PPL(CF(p))}{PPL(p)}.
\]
We report the mean over the central $95\%$ of values to reduce the influence of outliers.

\begin{table}[t]
	\centering
	\begin{adjustbox}{max width=\columnwidth}
		\input{tables/eval_result.tex}
	\end{adjustbox}
	\caption{Performance evaluation of \contentfuzz.\protect\footnotemark}
	\label{tab:eval}
\end{table}

\begin{figure*}[ht]
	\includegraphics[width=\textwidth]{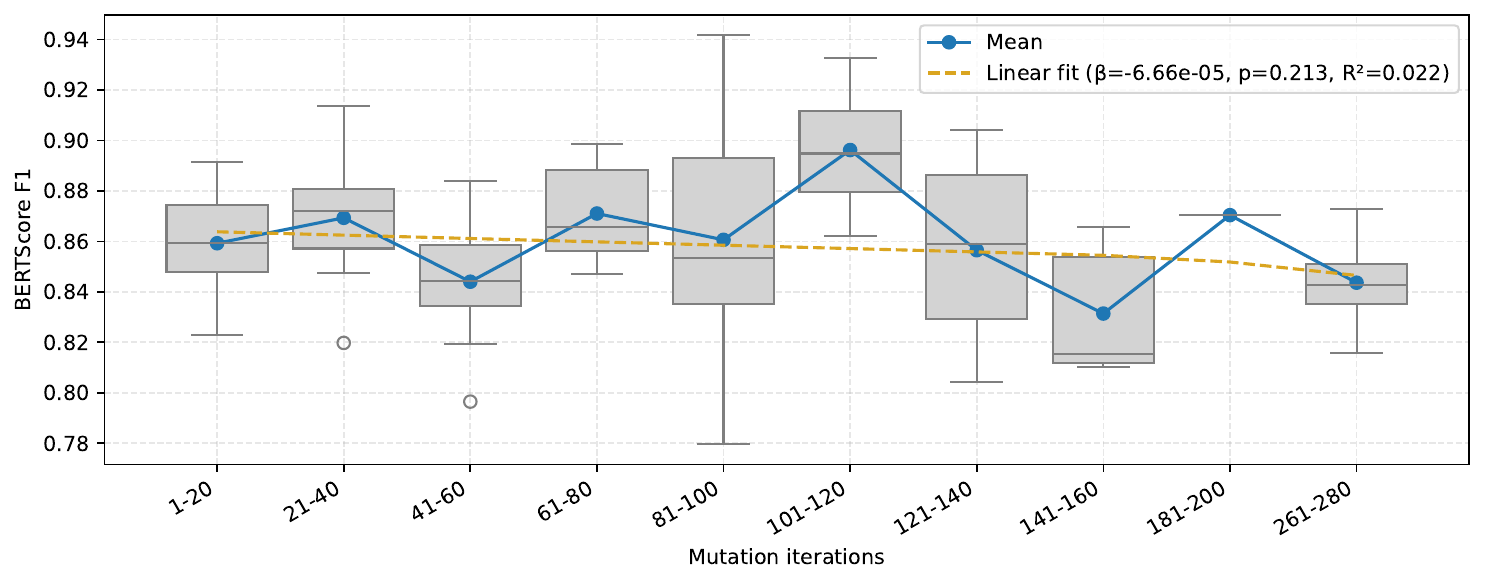}
	\caption{Semantic integrity over fuzzing iterations.}
	\label{fig:bertscore_vs_iter}
\end{figure*}

\footnotetext{Due to COLA running too slowly to obtain final results, we sampled 100 posts for this evaluation.}

Our evaluation results are summarized in \autoref{tab:eval}.
\contentfuzz with Gemini-2.5-Flash-Lite is effective across all targeted stance analyzers and datasets, achieving higher ESRs while maintaining strong semantic integrity and low perplexity.
Among the targeted stance analyzers, the zero-shot LLM analyzer is the most robust to \contentfuzz.
The ESRs on zero-shot analyzers are lower than those of the other analyzers, and the quality of the generated rewrites is also slightly lower.
Because the zero-shot analyzer is more robust,
successfully escaping posts require more aggressive rewrites that deviate further from typical language patterns,
resulting in higher absolute perplexity (\eg \num{112.634} on Sem16).
However, the perplexity ratio (PPLr) remains well below 1.0 (\eg \num{0.601}),
confirming that the rewrites are still fluent relative to their originals;
the high absolute PPL reflects the short, informal, and topically specific nature of the original Sem16 tweets.
We also observe that Chinese posts are easier to rewrite to escape information cocoons than English posts.
However, this comes at the cost of slightly lower semantic integrity.
We also compare against state-of-the-art adversarial attack methods in \appautoref{appendix:adv-comparison},
where we achieve $51\%$ relative improvement in success rate and over $90\%$ relative lower perplexity for generated rewrites.
We also provide case studies in \appautoref{appendix:case-study} to illustrate how \contentfuzz iteratively rewrites a post to change the prediction of the targeted analyzer.

\paragraph{NLI-based contradiction analysis}
BERTScore measures contextual similarity but cannot detect semantic inversions (\eg negation).
Natural language inference (NLI)~\cite{bowman2015snli} is the task of determining whether a hypothesis is entailed by, contradicts, or is neutral with respect to a given premise.
Following~\citet{kambhatla2024promoting}, who use NLI to verify meaning preservation in text rewriting,
we verify that rewrites do not contradict the originals, and in reverse.
We use \texttt{cross-encoder/nli-deberta-v3-large}~\cite{he2021debertav3} for English
and \texttt{MoritzLaurer/mDeBERTa-v3-base-mnli-xnli}~\cite{laurer2024less} for Chinese.
\autoref{tab:nli} summarizes the results.
The bidirectional entailment rate exceeds $90\%$ and the contradiction rate is $\leq 1.12\%$ across all datasets,
providing direct evidence that \contentfuzz rewrites preserve meaning
and complementing the embedding-based similarity captured by BERTScore with an explicit logical consistency check.

\begin{table}[t]
	\centering
	\small
		\input{tables/nli_result.tex}
	\caption{%
		NLI-based contradiction analysis on all successfully rewritten pairs. 
		The forward direction is from original to rewrite.
		Ent.: Entailment, Neu.: Neutral, Con.: Contradiction. 
		Values are in \%.
	}
	\label{tab:nli}
\end{table}

Finally, we analyze whether fuzzing progress reduces the semantic integrity of the generated rewrites.
\autoref{fig:bertscore_vs_iter} shows the semantic integrity of successfully rewritten posts over fuzzing iterations,
measured by BERTScore F1 on the Sem16 dataset with the fine-tuned RoBERTa.
We observe that the BERTScore remains relatively stable as the number of fuzzing iterations increases.
To quantify this trend,
we fit a linear regression between mean BERTScore and iteration index and find only a negligible negative coefficient
($\beta = -6.66\times10^{-5}$),
which is not statistically significant ($p = 0.213$) and explains little variance ($R^2 = 0.022$).
Indicated by the results,
we cannot conclude that more fuzzing iterations lead to systematic semantic degradation.

\subsection{Cross-model success rate} \label{sec:cross-model}

\begin{figure*}[t]
	\centering
	\includegraphics[width=\linewidth]{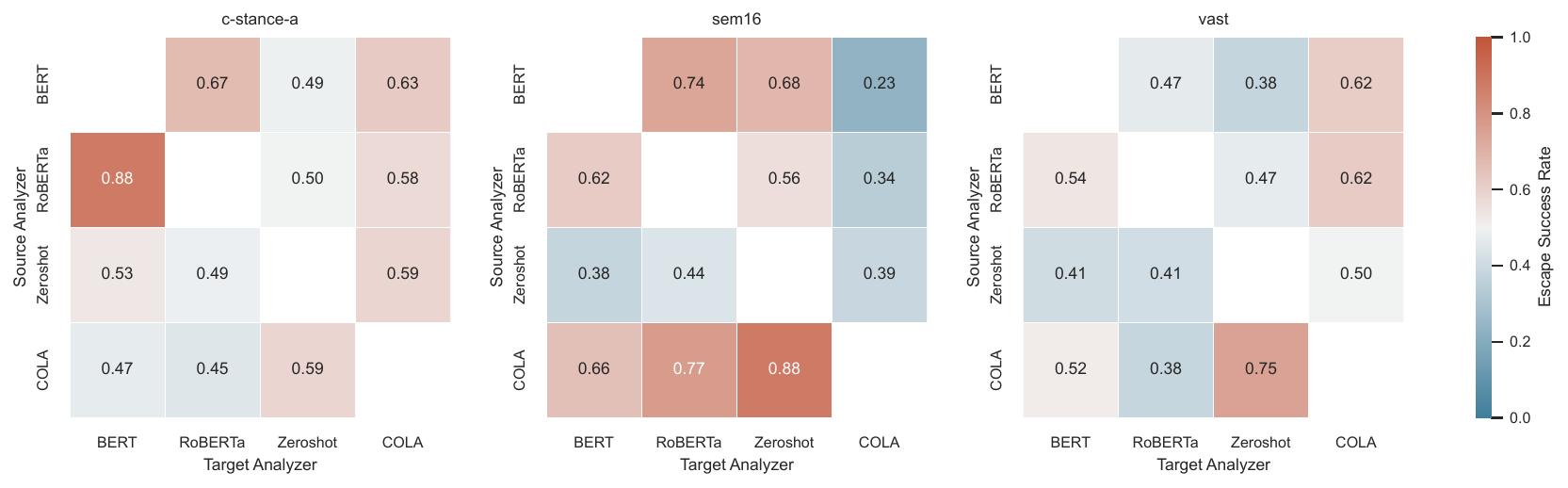}
	\caption{Cross-model transferability.}
	\label{fig:cross-model}
\end{figure*}

We investigate whether rewrites found for one targeted stance analyzer transfer to other unseen analyzers.
To this end,
we measure the extent to which escaping posts produced by fuzzing against one targeted stance analyzer
can also successfully escape other unseen stance analyzers~\cite{papernot2016transferability,liu2024autodan}.
For cross-model transferability, we take rewrites produced by fuzzing one target model and evaluate the misclassification rate on other unseen models, defined as $1-\mathrm{Acc}$,
where $\mathrm{Acc}$ is accuracy on those unseen models.

We present the cross-model transferability results in \autoref{fig:cross-model}.
We observe that models sharing the same architecture exhibit higher cross-model transferability.
For example, the fine-tuned BERT and RoBERTa demonstrate higher transferability with each other.
Furthermore, we find that COLA's cross-model success rate is very low for the Sem16 dataset,
but relatively high for the VAST and C-STANCE-A datasets.
We attribute this discrepancy to the fact that COLA uses manually designed expert roles for collaborative debates around the six topics in Sem16.
However, its performance and robustness do not generalize well to datasets with different topics and writing styles.
In addition, zero-shot LLM-based analyzers exhibit lower cross-model transferability than fine-tuned encoder-based models, indicating stronger robustness of LLMs against semantic-preserving rewrites.
Cross-model transferability is higher between architecturally similar models (\eg BERT $\leftrightarrow$ RoBERTa), while zero-shot LLM-based analyzers exhibit stronger robustness against transferred rewrites.

\subsection{Effects of temperature scheduling} \label{sec:ablation:temp-scheduling}

To analyze the effects of temperature scheduling,
we fix the seed scheduling strategy to \emph{priority queue} and fuzz the target stance analyzer with and without temperature scheduling.
For fuzzing without temperature scheduling,
we set the temperature to a constant value of 1.0,
which is the default value when accessing LLM APIs.
We report the mean, median, and standard deviation (std) of iterations required for successful posts.
We use RoBERTa as the targeted stance analyzer on the Sem16 dataset,
and the detailed settings are provided in \appautoref{appendix:temp_scheduling}.

\begin{table}[t]
	\centering
	\begin{adjustbox}{max width=\columnwidth}
		\input{tables/temp_sched.tex}
	\end{adjustbox}
	\caption{Effects of temperature scheduling.}
	\label{tab:temp_sched}
\end{table}

The statistics of the experiments are summarized in \autoref{tab:temp_sched}.
We observe a clear advantage of temperature scheduling over using a constant temperature of 1.0.
Since different posts have different wording, sentiments, styles, and stances,
fixing the temperature to a constant value limits the diversity of generated content,
and thus requires more fuzzing iterations to successfully rewrite the posts.
In contrast,
letting the fuzzer adapt the temperature during the fuzzing process
allows it to generate more diverse content,
which improves the efficiency and effectiveness of \contentfuzz.

\subsection{Effects of seed scheduling} \label{sec:ablation:seed-scheduling}

Modern fuzzers~\cite{fioraldi2020afl_pp,fioraldi2022libafl} support multiple seed scheduling strategies,
and users can choose different strategies based on their needs and their application domain.
To this end, we also designed multiple seed scheduling strategies for \contentfuzz
to accommodate different post topics on social media.
We implemented and evaluated four different seed scheduling strategies:
FIFO, random, weighted priority, and priority scheduling.
The priority scheduling strategy is described in \autoref{sec:seed-scheduling},
and the others are detailed in \appautoref{appendix:seed_scheduling}.
We follow the same settings as in \autoref{sec:ablation:temp-scheduling}.

\begin{table}[t]
	\centering
	\begin{adjustbox}{max width=\columnwidth}
		\input{tables/seed_sched.tex}
	\end{adjustbox}
	\caption{Effects of seed scheduling.}
	\label{tab:seed_sched}
\end{table}

From \autoref{tab:seed_sched},
we observe that the priority scheduling strategy outperforms all other strategies in ESR.
However, weighted probability scheduling is more efficient in terms of the maximum number of iterations required for successful posts,
with the lowest standard deviation as well.
This indicates that different seed scheduling strategies have different advantages.
We select the priority scheduling strategy as the default seed scheduling strategy for \contentfuzz,
since it achieves the highest ESR.

%% file: tables/eval_result.tex
\begin{tabular}{l S S S S}
	\toprule
	\textbf{Analyzer} & \textbf{ESR} & \textbf{BERTScore} & \textbf{PPL} & \textbf{PPLr} \\
	\midrule
	\rowcolor{lightgray} \multicolumn{5}{c}{\textit{Sem16}}                              \\
	BERT              & 0.563        & 0.889              & 71.335       & 0.422         \\
	RoBERTa           & 0.670        & 0.876              & 69.420       & 0.360         \\
	Zero-shot         & 0.773        & 0.885              & 112.634      & 0.601         \\
	COLA              & 0.480        & 0.882              & 52.247       & 0.458         \\
	\midrule
	\rowcolor{lightgray} \multicolumn{5}{c}{\textit{VAST}}                               \\
	BERT              & 0.883        & 0.878              & 10.041       & 0.322         \\
	RoBERTa           & 0.708        & 0.869              & 9.789        & 0.317         \\
	Zero-shot         & 0.655        & 0.892              & 24.448       & 0.749         \\
	COLA              & 0.410        & 0.896              & 16.787       & 0.537         \\
	\midrule
	\rowcolor{lightgray} \multicolumn{5}{c}{\textit{C-STANCE-A}}                         \\
	BERT              & 0.910        & 0.752              & 21.242       & 0.164         \\
	RoBERTa           & 0.879        & 0.774              & 16.717       & 0.163         \\
	Zero-shot         & 0.737        & 0.750              & 34.016       & 0.312         \\
	COLA              & 0.750        & 0.761              & 49.775       & 0.376         \\
	\bottomrule
\end{tabular}

%% file: tables/nli_result.tex
\begin{tabular}{c l r r r}
	\toprule
	\textbf{Direction}       & \textbf{Dataset} & \textbf{Ent.} & \textbf{Neu.} & \textbf{Con.} \\
	\midrule
	\mr{4}{$\longrightarrow$}   & Sem16            & 80.62                & 17.50                 & 1.88                     \\
	                        & VAST             & 97.56                & 2.28                  & 0.16                     \\
	                        & C-STANCE       & 94.03                & 5.38                  & 0.59                     \\
	                        & \textbf{All}     & \textbf{93.88}       & \textbf{5.54}         & \textbf{0.59}            \\
	\midrule
	\mr{4}{$\longleftarrow$}    & Sem16            & 66.67                & 29.06                 & 4.27                     \\
	                        & VAST             & 92.04                & 7.42                  & 0.54                     \\
	                        & C-STANCE       & 92.44                & 6.58                  & 0.99                     \\
	                        & \textbf{All}     & \textbf{90.46}       & \textbf{8.41}         & \textbf{1.12}            \\
	\bottomrule
\end{tabular}

%% file: tables/temp_sched.tex
\begin{tabular}{lcccc}
	\toprule
	\textbf{Temperature} & \textbf{ESR}   & \textbf{mean}   & \textbf{median} & \textbf{std}    \\
	\midrule
	1.0         & 0.620 & 16.702 & 2      & 36.362 \\
	Scheduling  & 0.670 & 15.324 & 2      & 36.931 \\
	\bottomrule
\end{tabular}

%% file: tables/seed_sched.tex
\begin{tabular}{lcccc}
	\toprule
	{\textbf{Scheduling}} & {\textbf{ESR}} & {\textbf{mean}} & {\textbf{median}} & {\textbf{std}} \\
	\midrule
	FIFO                  & 0.620          & 22.798          & 3                 & 47.830         \\
	Random                & 0.645          & 18.326          & 3                 & 35.513         \\
	Priority Random       & 0.665          & 15.985          & 2                 & 38.140         \\
	Priority              & 0.670          & 15.324          & 2                 & 36.931         \\
	\bottomrule
\end{tabular}

%% file: src/conclusion.tex
\section{Conclusion}

We present \contentfuzz, the first content-focused methodology that enables content creators to mitigate information cocoons on social media platforms.
\contentfuzz adopts a gray-box approach that leverages confidence scores from stance analyzers to guide an iterative rewriting process,
in which a generative LLM modifies post content.
The generated posts preserve the original, human-interpreted stance toward a given social topic,
while being classified differently by stance analyzers deployed on social media platforms.
\contentfuzz effectively generates diverse rewrites that escape information cocoons with high success rates,
while maintaining the original semantics of the posts.
We believe \contentfuzz represents a promising new direction in responsible AI for social media research,
with a particular focus on mitigating information cocoons.
Our source code is available at \url{https://github.com/EYH0602/ContentFuzz}.

%% file: src/limitations.tex
\section*{Limitations}

While \contentfuzz demonstrates promising results in mitigating information cocoons on social media platforms,
several limitations warrant consideration.
First, the current design of \contentfuzz focuses exclusively on stance detection,
which represents only one of the predictive components in modern recommender systems.
Future work could extend the methodology to additional predictors or to end-to-end recommender systems.
Second, we do not extensively optimize \contentfuzz or tune its hyperparameters to maximize success rates,
beyond designing temperature and seed scheduling strategies.
As \contentfuzz is the first work exploring content rewriting for escaping information cocoons,
our primary goal is to demonstrate feasibility rather than achieve optimal performance.
Third, \contentfuzz relies on confidence scores (\logprobs) from stance detection models as feedback to guide the rewriting process.
However, for some newer proprietary LLMs, these \logprobs are not directly accessible.
Fun-tuning~\cite{labunets2025funtuning} proposes estimating \logprobs using fine-tuning loss with a very small learning rate,
which could serve as an alternative.

Fourth, our evaluation relies on computational metrics rather than direct human annotation of meaning preservation.
While these metrics provide strong and complementary evidence,
they do not constitute a direct measurement of whether humans perceive the stance and intent as unchanged.
Nevertheless, our small-scale human evaluation (\appautoref{appendix:case-study}) confirms that the rewrites preserve the original meaning.
Finally, our evaluation is limited to empirical studies on public datasets and the aforementioned computational analysis metrics.
We do not examine downstream real-world impacts of deploying posts produced by \contentfuzz on production social media platforms
due to limited platform accessibility.

%% file: src/ethics.tex
\section*{Ethical considerations}

This work investigates how content creators may automatically rewrite posts to change the prediction of automated stance analyzers,
and thereby escape algorithmically induced information cocoons. 
The objective is to analyze and expose structural biases in stance-based recommender and moderation pipelines, 
rather than to facilitate deception, misinformation, or malicious manipulation. 
Accordingly, the rewriting process is strictly constrained to semantically preserving LLM-based rewrites that maintain the original intent and factual content of the post. 
\contentfuzz does not generate new content or introduce new claims; 
it rephrases existing content to probe the limitations of stance-based filtering mechanisms. 
All experiments are conducted on public datasets and models,
without targeting real users, platforms, or deployed production systems.
The case study examples are drawn and rewritten verbatim from publicly available research datasets~\cite{mohammad2016sem16,allaway2020vast,zhao2023c-stance};
user handles appearing in the original data have been anonymized to prevent identification.
These examples cover socially sensitive topics (e.g., abortion, feminism, religion) and are presented solely for research illustration;
their inclusion does not reflect the views of the authors.
While such techniques might be misused to evade automated moderation,
we frame our contribution as a diagnostic and exploratory study intended to improve transparency and robustness in stance-aware recommender systems.

%% file: src/appendix.tex
\appendix
\section{Appendix}

\subsection{Stance analyzer details}

\subsubsection{Fine-tuning} \label{appendix:fine-tuning}

We follow the provided split of each dataset for training, validation, and test.
For Sem16 and VAST,
we fine-tune the models with a learning rate of \num{2e-5}.
For C-STANCE-A,
we fine-tune the models with a learning rate of \num{5e-6}, following \citet{zhao2023c-stance}.
We fine-tune all models for 5 epochs with a batch size of 32
and select the checkpoint with the best validation macro-F1.
We conduct all fine-tuning on a single NVIDIA A100 40GB GPU.
For inference during fuzzing,
we use an NVIDIA GeForce RTX 3060 with 12GB memory.

\subsubsection{Prompts for stance analysis}

\input{fig/prompt_cls.tex}
\begin{table}[h]
	\centering
	\begin{adjustbox}{max width=\columnwidth}
		\input{tables/stance_perforance.tex}
	\end{adjustbox}
	\caption{Performance of different stance analyzers.}
	\label{tab:performance}
\end{table}

\subsection{Prompts for content mutation}

\input{fig/prompt_en.tex}
\input{fig/prompt_zh.tex}

\subsection{Experiment settings and details}

\subsubsection{Temperature scheduling} \label{appendix:temp_scheduling}

In this subsection,
we evaluate the effectiveness of temperature scheduling and other components of \contentfuzz.
Specifically,
we analyze their effects from two perspectives:
\begin{enumerate}
	\item Performance: We follow \autoref{sec:performance-evaluation} to measure the escape success rate (ESR) of \contentfuzz under different configurations.
	\item Resource efficiency: We report the distribution of the number of iterations that \contentfuzz needs to rewrite posts.
\end{enumerate}
We perform all ablation studies on 200 tasks randomly sampled from the Sem16 dataset~\cite{mohammad2016sem16}.
We fix the maximum number of iterations at 300 for all experiments.
We use the fine-tuned RoBERTa~\cite{liu2019roberta} from \autoref{sec:stance-analyzers} as the targeted stance analyzer.

\subsubsection{Seed scheduling} \label{appendix:seed_scheduling}

\paragraph{First-in-first-out (FIFO)}
We implement a simple FIFO queue to store the seed posts as a baseline scheduling strategy,
\ie, without scheduling.
When the fuzzer considers a seed post interesting,
it appends the post to the end of the queue.
To select the next seed post to fuzz,
the fuzzer takes the seed post at the front of the queue.

\paragraph{Random}
The random seed scheduling strategy samples a seed from the seed pool in each iteration.
It assigns each seed in the pool equal probability,
regardless of their previous fuzzing results.

\paragraph{Weighted}
The weighted seed scheduling strategy assigns different weights to different seed posts in the seed pool.
We compute the weights from the confidence scores of the targeted stance analyzer on the seed posts.
Let $s$ denote a seed post in the seed pool, and let $W(\cdot)$ denote its weight,
\[
	W(s) = \frac{1}{\conf(s.content, s.stance)}.
\]
Then the probability $P(\cdot)$ of picking the seed $s$ from the seed pool is
\[
	P(s) = \frac{W(s)}{\sum_{s' \in \text{seed pool}} W(s')}.
\]
The strategy samples seeds with lower confidence scores more often,
but it can still pick any seed by chance.

\subsection{Comparison with adversarial methods} \label{appendix:adv-comparison}

\input{src/adversarial.tex}

\subsection{Case study} \label{appendix:case-study}

\begin{table*}[ht]
	\centering
	\small
	\input{tables/case_study.tex}
    \caption{Case study illustrating confidence-guided content fuzzing across iterations. Topic: Atheism.}
    \label{tab:case_study_example}
\end{table*}

We present a case study in \autoref{tab:case_study_example} to illustrate how \contentfuzz iteratively rewrites a post to change the prediction of the targeted stance analyzer.
Our case is sampled from the Sem16 dataset~\cite{mohammad2016sem16} with the topic of Atheism.
Starting from the original post in iteration 0,
\contentfuzz gradually rewrites the post across iterations 1 and 2.
The targeted stance analyzer initially predicts the original post as \texttt{Against} in iteration 0.
After the first iteration of rewriting,
the fuzzer generates a new post that is semantically equivalent to the original but with a
lower confidence score.
Using this new post as the seed,
\contentfuzz further rewrites the post in iteration 2,
which successfully flips the predicted stance to \texttt{Favor},
while maintaining the original meaning of the post.

\autoref{tab:case_study_extended} presents additional examples across all three datasets and both languages.
The \emph{Stance Change} column shows the ground-truth label and the analyzer's (incorrect) prediction (\textcolor{BrickRed}{red}).
These examples illustrate several recurring patterns:

\paragraph{Surface register \vs underlying stance}
In the Sem16 Feminist Movement and Legalization of Abortion examples,
the rewrites soften the tone or formalize the register while preserving the core argument.
For example, the Feminist Movement rewrite adopts a more measured register but advances the same pro-feminist argument,
yet the model flips from \texttt{Favor} to \texttt{Against} with high confidence.
Similarly, the Legalization of Abortion rewrite retains the original anti-abortion position in calmer phrasing,
and BERT loses the \texttt{Against} signal.

\paragraph{Factually identical arguments}
The VAST 3D Printing example shows that even when the rewrite preserves every factual claim (the same technical limitations in material variety and speed), BERT reverses its prediction from \texttt{Against} to \texttt{Favor}.
This suggests that encoder-based models attend to phrasing cues (e.g., hedging constructions) rather than propositional content.

\paragraph{Zero-shot analyzer robustness}
The VAST NATO example shows that LLM-based zero-shot analyzers are not immune:
despite the rewrite retaining an explicit call to dissolve NATO,
the analyzer shifts to \texttt{Neutral} after 13 iterations.
The higher iteration count is consistent with the lower ESR of zero-shot analyzers reported in \autoref{sec:performance-evaluation}.

\paragraph{Cross-lingual transferability}
The C-STANCE Conservative Groups example extends the case study to Chinese.
The rewrite maintains the same favorable position on conservative groups,
yet Chinese RoBERTa flips from \texttt{Favor} to \texttt{Neutral} with 0.98 confidence.
This suggests that paraphrase-based content fuzzing generalizes across languages.

\paragraph{Human verification}
To complement the computational metrics reported in the main paper,
the authors and two independent PhD-level researchers from different domains independently reviewed all case study examples in \autoref{tab:case_study_example} and \autoref{tab:case_study_extended}.
All reviewers agreed that the rewritten posts preserve the original core meaning,
confirming that the semantic changes introduced by \contentfuzz are superficial rather than substantive.

\begin{table*}[h]
	\centering
	\begin{adjustbox}{max width=\textwidth}
	\input{tables/case_study_extended.tex}
	\end{adjustbox}
    \caption{Case study examples across datasets, analyzers, and languages.
    The first column lists the dataset and target analyzer; the second column lists
    the topic, stance change (ground-truth $\to$ analyzer's incorrect prediction in \textcolor{BrickRed}{red}),
    and the analyzer's confidence on the fuzzed post.}
    \label{tab:case_study_extended}
\end{table*}

%% file: fig/prompt_cls.tex
\begin{figure}[h]
    \centering
    \begin{tcolorbox}[title=System Instruction]
        \scriptsize\ttfamily
        You are a precise stance classifier.
        Decide whether the author's attitude is Favor / Against / Neutral towards the target {target}.
        Be conservative: if unclear, choose Neutral.
        ONLY output one word chosen from Favor, Against, Neutral.
    \end{tcolorbox}
    \caption{The system instruction for generative stance analysis (zero-shot and prompt-engineering).}
    \label{fig:cls_prompt}
\end{figure}

%% file: tables/stance_perforance.tex
\begin{tabular}{
		l
		cc
		cc
		cc
	}
	\toprule
	\textbf{Model}
	 & \multicolumn{2}{c}{\textbf{C-STANCE-A}}
	 & \multicolumn{2}{c}{\textbf{Sem16}}
	 & \multicolumn{2}{c}{\textbf{VAST}}                         \\
	 & {\textit{Acc}}                          & {\textit{F1}}
	 & {\textit{Acc}}                          & {\textit{F1}}
	 & {\textit{Acc}}                          & {\textit{F1}}   \\
	\midrule

	BERT
	 & 0.76                                    & 0.76
	 & 0.62                                    & 0.53
	 & 0.70                                    & 0.70          \\

	RoBERTa
	 & 0.78                                    & 0.78
	 & 0.65                                    & 0.62
	 & 0.74                                    & 0.73          \\
	\lightmidrule

	Zero-shot %
	 & 0.52                                    & 0.52
	 & 0.58                                    & 0.56
	 & 0.57                                    & 0.56          \\

	\lightmidrule

	COLA
	 & 0.49                                    & 0.41
	 & 0.67                                    & 0.29
	 & 0.41                                    & 0.31              \\

	\bottomrule
\end{tabular}

%% file: fig/prompt_en.tex
\begin{figure}[h]
    \centering
    \begin{tcolorbox}[title=System Instruction]
        \scriptsize\ttfamily
        You are a helpful writing assistant and an avid social media user.
        Your role is to help the content creator refine their post to make it more engaging and shareable.
        Improve the writing and flow while **keeping the post's original meaning intact**.
        **Do not change the author's stance** (their position or opinion on the topic) **or the target topic** of the post.
        Make sure to **preserve the original tone, style, and sentiment** of the writing, maintaining the author's voice.
        Only **make minimal edits**: the goal is to polish the text, not to overhaul it.
        Output **only** the revised text, and do not include any explanations.
        Always **keep the content in the same language** as the original post (no translation or dialect change).
        Do not extensively use emojis or hashtags unless they were present in the original text.
    \end{tcolorbox}


    \begin{tcolorbox}[title=Prompt Template]
        \scriptsize\ttfamily
        The current text is \{stance\} towards the \{target\}.
        Without changing its meaning, please rewrite the following text:\\
        ```\\
        \{text\}\\
        ```
    \end{tcolorbox}
    
    \caption{The system instruction and prompt template for LLM-based rewrite mutation.}
    \label{fig:rewrite_prompts_en}
\end{figure}

%% file: fig/prompt_zh.tex
\begin{figure}[h]
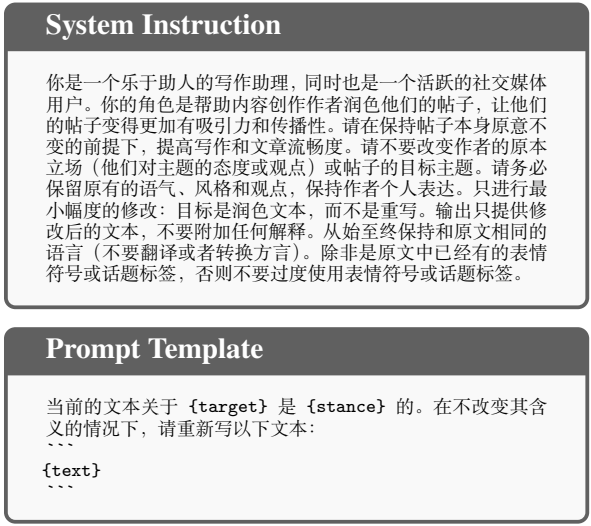

    \centering
    \begin{tcolorbox}[title=System Instruction]
        \scriptsize\ttfamily
        你是一个乐于助人的写作助理，同时也是一个活跃的社交媒体用户。
        你的角色是帮助内容创作作者润色他们的帖子，让他们的帖子变得更加有吸引力和传播性。
        请在保持帖子本身原意不变的前提下，提高写作和文章流畅度。
        请不要改变作者的原本立场（他们对主题的态度或观点）或帖子的目标主题。
        请务必保留原有的语气、风格和观点，保持作者个人表达。
        只进行最小幅度的修改：目标是润色文本，而不是重写。
        输出只提供修改后的文本，不要附加任何解释。
        从始至终保持和原文相同的语言（不要翻译或者转换方言）。
        除非是原文中已经有的表情符号或话题标签，否则不要过度使用表情符号或话题标签。
    \end{tcolorbox}


    \begin{tcolorbox}[title=Prompt Template]
        \scriptsize\ttfamily
        当前的文本关于\{target\}是\{stance\}的。
        在不改变其含义的情况下，请重新写以下文本：\\
        ```\\
        \{text\}\\
        ```
    \end{tcolorbox}
    
    \caption{The Chinese system instruction and prompt template for LLM-based rewrite mutation.}
    \label{fig:rewrite_prompts_zh}
\end{figure}

%% file: src/adversarial.tex
Another line of related work concerns adversarial attacks~\cite{meng2017MagNet,li2020nobox,Jia2020Fooling,guo2020backpropagating,li2025bayesian}.
Adversarial attacks on classification models add noise to the original inputs to mislead the model, producing adversarial examples.
Attackers often optimize these perturbations to be imperceptible to humans.
For text classification, adversarial examples typically preserve semantics.
Although our work does not aim to attack stance analyzers or recommender systems,
our content mutation task shares characteristics with adversarial attacks.
We compare \contentfuzz with two state-of-the-art adversarial attack methods.

\subsubsection{Baselines and experimental settings}

\paragraph{BERT-Attack}
BERT-Attack~\cite{li2020bert-attack} targets text classification models fine-tuned on BERT.
BERT-Attack first identifies vulnerable words in the input text that most influence the model's prediction,
then iteratively replaces the ranked words with BERT-based lexical substitutions~\cite{zhou2019bert-based}.
BERT-Attack aims to minimize the perturbation rate while achieving a high success rate.
We use the officially released code\footnote{\url{https://github.com/LinyangLee/BERT-Attack}}
in our evaluation.

\paragraph{Reinforce-Attack}
\citet{gao2024semantic} proposed Reinforce-Attack,
which generates semantic-preserving adversarial examples against BERT-based classifiers.
Reinforce-Attack utilizes a reinforcement learning framework to optimize the generation of adversarial examples,
where the attack process is controlled by a reward function rather than heuristic rules.
The reward function encourages higher semantic similarity and lower query costs,
and the method achieves significantly higher semantic similarity than BERT-Attack
while maintaining comparable attack success rates.
Because the authors did not release code,
we reimplement Reinforce-Attack based on the descriptions in the original paper.

We evaluate these methods on the Sem16, VAST, and C-STANCE-A datasets with a fine-tuned BERT stance analyzer
because they do not support other model architectures or languages.
To the best of our knowledge, no existing adversarial attack methods can be generalized to encoder-based,
zero-shot generative, and prompt-engineering-based stance analyzers as \contentfuzz does.

\subsubsection{Result analysis}

\begin{table}[t]
	\centering
	\begin{adjustbox}{max width=\columnwidth}
		\input{tables/adv_result.tex}
	\end{adjustbox}
	\caption{Comparison between \contentfuzz and adversarial attack methods on stance detection.}
	\label{tab:adv_result}
\end{table}

We present the comparison results in \autoref{tab:adv_result}.
\contentfuzz outperforms BERT-Attack and Reinforce-Attack by a large margin in attack success rate (ASR) and fluency (PPL and PPLr) across all three datasets.
Although the BERTScore of \contentfuzz is slightly lower than that of BERT-Attack and Reinforce-Attack,
it remains within an acceptable range.
This difference arises because BERTScore is computed from the similarity between embeddings of tokenized text.
Unlike \contentfuzz, which rewrites the text,
these adversarial attack methods preserve the positions of most tokens,
a property that BERTScore favors because of positional encoding.
However, because the substituted tokens are often nonsensical,
the fluency of the generated adversarial examples degrades substantially,
as indicated by their high perplexity.
Overall, \contentfuzz demonstrates superior performance in generating effective and fluent content mutations,
even when considered as a form of adversarial attack.

%% file: tables/adv_result.tex
\begin{tabular}{l S S S S}
	\toprule
	\textbf{Analyzer} & \textbf{ASR} & \textbf{BERTScore} & \textbf{PPL} & \textbf{PPLr} \\
	\midrule
	\rowcolor{lightgray} \multicolumn{5}{c}{\textit{Sem16}}                              \\
	BERT-Attack       & 0.371        & 0.934              & 1246.836     & 4.119         \\
	Reinforce-Attack  & 0.177        & 0.970              & 464.794      & 1.613         \\
	\contentfuzz      & 0.563        & 0.889              & 71.335       & 0.422         \\
	\midrule
	\rowcolor{lightgray} \multicolumn{5}{c}{\textit{VAST}}                               \\
	BERT-Attack       & 0.679        & 0.959              & 81.2616      & 2.3789        \\
	Reinforce-Attack  & 0.191        & 0.995              & 38.772       & 1.055         \\
	\contentfuzz      & 0.883        & 0.878              & 10.041       & 0.322         \\
	\midrule
	\rowcolor{lightgray} \multicolumn{5}{c}{\textit{C-STANCE-A}}                         \\
	Reinforce-Attack  & 0.003        & 0.916              & 886.696      & 2.758         \\
	\contentfuzz      & 0.910        & 0.752              & 21.242       & 0.164         \\
	\bottomrule
\end{tabular}

%% file: tables/case_study.tex
\begin{tabular}{c l c p{7.2cm}}
\toprule
\textbf{Iteration} & \textbf{Stance} & \textbf{Confidence} & \textbf{Post} \\
\midrule
0 & Against & 1.0000 &
I am human. I look forward to the extinction of humanity with eager anticipation. We deserve nothing less. \\
\lightmidrule
1 & Against & 0.4582 &
I am human, and I eagerly await humanity's extinction. It's what we deserve. \\
\lightmidrule
2 & Favor & 0.4496 &
I am human, and I cannot wait for humanity's extinction. It's what we deserve. \\
\bottomrule
\end{tabular}

%% file: tables/case_study_extended.tex
\setlength{\tabcolsep}{4pt}
\begin{tabularx}{1.3\textwidth}{p{2.3cm} p{3.4cm} X X}
\toprule
\textbf{Dataset} & \textbf{Stance Change} & \textbf{Original Post} & \textbf{Fuzzed Post} \\
\midrule
Sem16 \newline {\small\textit{RoBERTa}} &
{\textit{Feminist Movement}} \newline Favor $\to$ \textcolor{BrickRed}{Against} \newline Conf.\ 0.86 &
Expel them\_ male millionaires from society! All Rights for women and children! Stop children women trafficking! &
Let's champion the Feminist Movement by advocating for the rights of women and children, and by putting an end to the trafficking of vulnerable individuals. \\
\midrule
Sem16 \newline {\small\textit{BERT}} &
{\small\textit{Legalization of Abortion}} \newline Against $\to$ \textcolor{BrickRed}{Neutral} \newline Conf.\ 0.67 &
@user1 @user2 @user3 Actually, child-murder is far worse these days. We live in more savage times. &
In these difficult times, we are witnessing a disturbing increase in the loss of young lives. \\
\midrule
VAST \newline {\small\textit{BERT}} &
{\small\textit{3D Printing}} \newline Against $\to$ \textcolor{BrickRed}{Favor} \newline Conf.\ 0.76 &
I can see 3D printing for prototypes, and some custom work. However manufacturing industries use thousands of plastics and thousands of metal alloys, few of which can be printed. I don't see that there is going to be a wholesale conversion to this relatively slow, materials-inflexible process. &
While 3D printing is useful for prototypes and custom items, its broad use in manufacturing is still facing major challenges. The industry utilizes a huge range of plastics and metal alloys, but only a small percentage can currently be used with 3D printing. It's doubtful that this process will completely replace current methods due to its limitations in speed and material options. \\
\midrule
VAST \newline {\small\textit{Gemini}} &
{\small\textit{NATO}} \newline Against $\to$ \textcolor{BrickRed}{Neutral} \newline Conf.\ 0.67 &
If we need to develop a world police force with other nations than we should have that conversation as to the how, the why and the cost. In the meantime it is long past time to draw NATO to a close. &
Before we entertain the idea of a global police force, a deep dive into its practicalities, objectives, and costs is essential. Meanwhile, dissolving NATO is long overdue. \\
\midrule
C-STANCE \newline {\small\textit{RoBERTa}} &
{\small\textit{Conservative Groups}} \newline Favor $\to$ \textcolor{BrickRed}{Neutral} \newline Conf.\ 0.98 &
浅析一下保守群体，其实不是一概而论。有些自由意志主义者，奉行不干预市场，但在社会价值观上允许堕胎和LGBTQ。有些新保守主义者支持军事干预外部事务。而真保守派，持守新教信仰反堕胎反性别错乱，对外奉行不干预的孤立主义，支持低税不干预市场，强调社会秩序打击犯罪。 &
浅析保守群体，实际上并非铁板一块。比如，一些自由意志主义者信奉自由市场，却在堕胎和LGBTQ议题上持开放态度。另一些新保守主义者则倾向于在国际事务中采取军事干预。而传统保守派，则坚守新教信仰，反对堕胎和性别多元化，对外奉行不干预的孤立主义，支持低税收和自由市场，并着重强调社会秩序与打击犯罪。 \\
\bottomrule
\end{tabularx}